%% file: paper.tex
\documentclass{article}
\usepackage[ruled]{algorithm}
\usepackage{algpseudocode}
\usepackage{graphicx}
\usepackage{subcaption}
\usepackage{amsmath}
\usepackage{amsfonts}
\usepackage{amssymb}
\usepackage{amsthm}
\usepackage{mathtools}
\usepackage{booktabs}
\usepackage{wrapfig}
\usepackage{hyperref}
\usepackage{tikz}
\usetikzlibrary{positioning, shapes.geometric, decorations.pathreplacing}

\title{Fast Neural Inverse Kinematics\\ on Human Body Motions}

\author{David Tolpin,\\Yoom\\ davidt@yoom.com \and Sefy Kagarlitsky, \\Yoom\\ sefy@yoom.com}

\begin{document}

\maketitle
\begin{abstract}
Markerless motion capture enables the tracking of human motion
without requiring physical markers or suits, offering increased
flexibility and reduced costs compared to traditional systems.
However, these advantages often come at the expense of higher
computational demands and slower inference, limiting their
applicability in real-time scenarios. In this technical report,
we present a fast and reliable neural inverse kinematics
framework designed for real-time capture of human body motions from 3D
keypoints. We describe the network architecture, training
methodology, and inference procedure in detail. Our framework is
evaluated both qualitatively and quantitatively, and we support
key design decisions through ablation studies.
\end{abstract}

\section{Introduction}
\label{sec:introduction}

Markerless motion capture seeks to infer human motion directly
from visual input, eliminating the need for physical markers or
specialized suits. This approach has gained traction in recent
years due to its lower cost, greater flexibility, and ease of
deployment compared to traditional marker-based systems. It has
become particularly valuable in domains such as animation,
gaming, sports science, virtual reality (VR), and human–computer
interaction (HCI). However, the accurate and efficient
reconstruction of full-body motion sequences—--not just individual
poses--—remains a major challenge, particularly when real-time
performance is required.

Many recent systems employ a three-stage pipeline: first
extracting 2D keypoints using pose estimation models such
as OpenPose~\cite{cao2019openpose} or
Detectron2~\cite{wu2019detectron2}, then triangulating them to receive 3D keypoints, and finally applying inverse
kinematics (IK) to estimate the underlying joint rotations and
global translations. In this work we focus on the third stage.

While traditional IK methods can be applied
to individual poses, recovering plausible temporally coherent
motion from sequences of keypoints requires modeling both
spatial and temporal dependencies. Optimization-based IK
approaches~\cite{bogo2016keep} often provide accurate motion
reconstructions but are computationally expensive and too slow
for real-time use.  Similarly, sequential models such as RNNs or
temporal convolutions~\cite{pavllo2019video} may capture
temporal structure but often struggle with long-range
dependencies or do not meet performance requirements.

To address these challenges, we propose a fast and reliable
neural IK framework tailored for real-time reconstruction of
human body motions from 3D keypoint sequences. Our model takes as
input a sequence of 3D keypoint detections and predicts corresponding body shape, joint rotations and global root translations for each frame. Crucially, our method does not treat frames independently; it explicitly models the dynamics of motion over time.

At the core of our method is a transformer-based encoder.
Transformers have proven effective in human motion modeling
tasks, including motion prediction~\cite{mao2019learning} and
generation~\cite{aksan2021spatio}, due to their ability to
attend over long temporal windows and represent complex
dependencies. We adapt these strengths to the inverse kinematics
setting, enabling our model to produce physically consistent and
temporally smooth motion reconstructions in real time.

Our system is trained end-to-end on our proprietary motion dataset.
Unlike per-frame regression models, our architecture exploits temporal
context to disambiguate noisy keypoints and to produce
kinematically plausible outputs across entire motion sequences.
We evaluate the framework qualitatively and quantitatively, and
perform ablation studies to justify our design choices.

In summary, our contributions are as follows:

\begin{itemize}
\item We propose a transformer-based inverse kinematics
framework designed specifically for fast and accurate
reconstruction of full-body motion sequences from 3D keypoints.

\item Our model performs real-time inference of joint
rotations and global translations, leveraging temporal context
for improved stability and realism.

\item We describe the architecture and training procedure in
detail, and provide empirical evaluations including ablations to
support our design.
\end{itemize}


\section{Dataset}
\label{sec:dataset}
Our proprietary motion dataset is the result of years of dedicated effort in capturing and processing high-quality volumetric video data using Yoom's volumetric system. Collected in-house, it represents a diverse range of human motions recorded in controlled environments. Each volumetric video sequence is topologically aligned using a registration algorithm that leverages both geometry and texture. Subsequently, the data is processed by fitting a parametric human body model to the footage using an optimization technique that, while time-consuming, yields highly accurate representations of body pose and shape.

This rigorous process has produced a dataset which is organized as a list of motion sequences, each representing continuous human motion of a subject. Every sequence includes a fixed body shape represented by a vector and a series of frames. Each frame contains joint rotations, representing the pose of the body, and a global root translation, indicating the position of the body in 3D space.

For the purpose of composing the training set to train our system, we extracted 3D keypoint from our dataset using an approach similar to the method of~\cite{kanazawa2018end}, where we linearly mapped  predefined vertices of our human body model to BODY25~\cite{cao2019openpose} keypoint set.

\section{Network, Training, and Inference}
\label{sec:network}

In order to facilitate fast markerless motion capture, we need
to decide on 
\begin{itemize}
\item the neural network architecture, including input and
output representations;
\item the training data, its transformation and augmentation;
\item the training objectives and network regularization;
\item the inference algorithm, which, given a trained model and
new data, returns a parameterized motion.
\end{itemize}

In what follows, we assume that reliable, if noisy, 3D keypoints
are obtained for each frame of the motion using existing methods~\cite{amin2013multi},~\cite{dong2019fast}~\cite{liao2024multiple}. These methods are well established, and their leverage allows
us to concentrate on the most critical stage of markerless
motion capture: conversion from a sequence of 3D keypoints to
translations, joint rotations, and body shape parameters.

\subsection{Network Architecture}

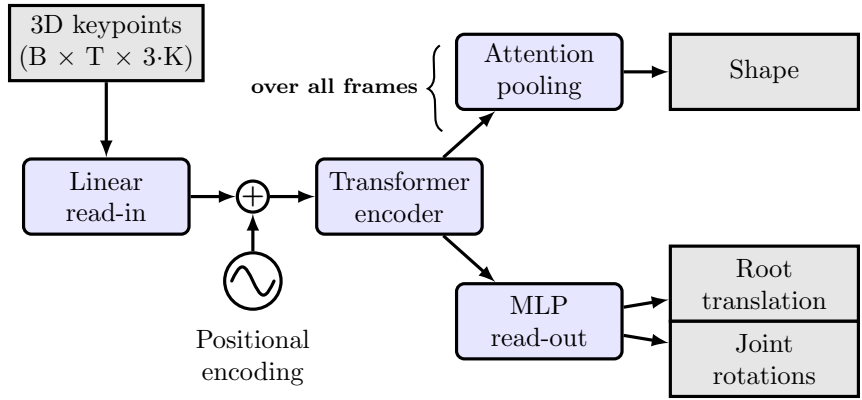
\begin{figure}
\input{network}
\caption{Network architecture}
\label{fig:network}
\end{figure}

Our network architecture is shown in
Figure~\ref{fig:network}. The input to the model is a motion
sequence, represented as a series of 3D keypoints over time---for
example, the 2D BODY25 keypoints output by OpenPose, triangulated to 3D. These
keypoints are first augmented with additive positional encodings
and then passed through a linear read-in layer to project them
into the internal feature space of the transformer.

The transformer encoder processes the entire motion sequence,
capturing both spatial and temporal dependencies. Its output is
then split into two prediction branches: one estimates per-frame
the global root translations and joint rotations; the other
estimates global shape parameters that are shared across all
frames. We represent joint rotations
using a 6D continuous representation~\cite{Zhou_2019_CVPR} for improved stability
during training.

While the architecture may appear simple, it incorporates key
design principles of neural inverse kinematics. The read-in
layer prepares the 3D keypoints for temporal modeling. The
transformer backbone models inter-frame dependencies while also
providing sufficient capacity for inferring plausible kinematics
at each frame. Finally, a temporal attention pooling mechanism
aggregates information across time to predict shape parameters,
assigning greater weight to frames that are more informative for
estimating body shape.

\subsection{Training Data}

Our motion dataset consists of motion sequences of varying
durations and frame rates. For training, we downsample all
sequences to the same frame rate. To produce uniformly sized
training samples, each motion is divided into fixed-length
temporal chunks. A common strategy is to apply a sliding window
with a stride equal to half the chunk length. Final segments
shorter than half the chunk length are discarded, while those
longer than half but shorter than the full length are padded by
repeating their last frame until the desired length is reached. 

Absolute location and orientation of a motion are rather
arbitrary. In many datasets there are just a few orientations
of motion directions, probably influenced by the capturing system setup.
As an augmentation, we rotate each motion around the Z axis (which is up in our coordinate system) by a
random angle sampled uniformly from $[0, 2\pi)$. We do not
modify the motion's location, but formulate the translation loss
in a location-invariant way (see Section~\ref{sec:losses}).

We standardize the keypoints such that each frame's keypoints
have zero mean along each axis and unit standard deviation over
all axes, combined.  Some inverse kinematics models add
synthetic noise to the keypoints during training to simulate
real-world uncertainty. We did not find this necessary: our
model performs well even when trained on clean, noise-free
keypoints from our dataset.

\subsection{Losses}
\label{sec:losses}

The model is trained to predict 
\begin{enumerate}
\item joint rotations;
\item root translation;
\item body shape parameters.
\end{enumerate}

\paragraph{Rotations} The loss on predicted rotations is
rather straightforward, and should penalize for the discrepancy
between the predicted and the true rotations. We impose the
geodesic loss: first, 6D rotations are transformed into rotation
matrices; then, the geodesic loss $\mathcal{L}_g$ between the
predicted $R_p$ and the true $R_t$ rotation is computed as 

\begin{equation}
\mathcal{L}_g = \arccos \left(\frac {\mathrm{tr}(R_pR_t^T) - 1}  2 \right)
\label{eqn:geodesic-loss}
\end{equation}

We regularize the 6D rotations by imposing the orthonormality
loss $\mathcal{L}_{on}$.  Given a 6D rotation representation
$R = [\mathbf{a}, \mathbf{b}] \in \mathbb{R}^{6}$, the
orthonormality loss is defined as:
\begin{equation}
\mathcal{L}_{on} = \left( \|\mathbf{a}\|^2 - 1 \right)^2 + \left( \|\mathbf{b}\|^2 - 1 \right)^2 + \frac{(\mathbf{a}^\top \mathbf{b})^2}{\|\mathbf{a}\|^2 \cdot \|\mathbf{b}\|^2}
\label{eqn:orthonormality-loss}
\end{equation}
This loss is low-weight, designed to prevent severe deviations from unit norm and orthogonality between the two 3D vectors.

\paragraph{Shape and translations} It is seemingly
straightforward to penalize for the discrepancy between the
predicted and the true shape and translation. However, we
center and standardize keypoints on input to avoid overfitting and
promote generalization, hence we cannot predict the translation
directly. Instead, we use a variant of \textit{cycle
consistency} loss to learn to predict both the shape vector and
the root translations. After predicting the shape parameters,
the root translations, and the joint rotations, we generate the
predicted keypoints $K_p$ via the regressor just as we did for
the true keypoints $K_t$ (see Section~\ref{sec:dataset}), and standardize \textit{but do not
center} them. Then, the loss is the mean squared difference
between $K_p$ and $K_t$:
\begin{equation}
\mathcal{L}_{cc} = \mathrm{mse}(K_p, K_t)
\label{eqn:cycle-loss}
\end{equation}
With this loss, the predicted translation is relative to the
center of mass (the mean) of the input keypoints.

The overall training loss is the sum of all the three losses
introduced above:
\begin{equation}
\mathcal{L} = \mathcal{L}_g + \mathcal{L}_{on} + \mathcal{L}_{cc}
\label{eqn:loss}
\end{equation}

\subsection{Inference}

Once the model is trained, inference reduces to passing a
sequence of 3D pose keypoints through the network and reading
off the estimated model parameters. However, because we target
real-time (online) applications and the model is trained on
motion chunks of fixed length $L$, several design decisions must
be made regarding how incoming frames are processed and how
outputs are interpreted.

The simplest approach is to maintain a sliding window of the $L$
most recent frames (or fewer, during warmup), pass the current
chunk to the model, and extract the pose and translation
parameters for the last frame in the window. This scheme,
illustrated in Algorithm~\ref{alg:one-by-one}, allows
immediate per-frame inference with zero delay.

\begin{algorithm}
\caption{Inference: one-by-one}
\label{alg:one-by-one}
\begin{algorithmic}[1]
	\Require model $f$, window size $L$
	\State Initialize buffer $X \gets [\,]$
	\For{each new frame with keypoints $x_t$}
		\State Append $x_t$ to $X$
		\If{length of $X > L$} \State Remove oldest frame \EndIf
	    \State $\hat{\Theta} \gets f(X)$
		\State Output $\hat{\Theta}_{t}[L - 1]$
	\EndFor
\end{algorithmic}
\end{algorithm}

While this strategy yields minimal latency, it tends to produce
less accurate or noisier predictions near chunk boundaries.
Since central frames in the input window benefit more from
temporal context, one can instead allow a small delay of $d < L$
frames and always extract the estimate corresponding to the $L
- d$-th frame in the window (Algorithm~\ref{alg:lookahead}).
This improves accuracy at the cost of a fixed delay.

\begin{algorithm}
\caption{Inference: lookahead}
\label{alg:lookahead}
\begin{algorithmic}[1]
\Require model $f$, window size $L$, delay $d$
\State Initialize buffer $X \gets [\,]$
\State Initialize output queue $Q \gets [\,]$
\For{each new frame with keypoints $x_t$}
	\State Append $x_t$ to $X$
	\If{length of $X > L$} \State Remove oldest frame \EndIf
	\If{length of $X = L$}
		\State $\hat{\Theta} \gets f(X)$
		\State Append $\hat{\Theta}[L-d - 1]$ to $Q$
	\EndIf
	\If{$Q$ not empty}
	\State Output and remove first element from $Q$
	\EndIf
\EndFor
\end{algorithmic}
\end{algorithm}

Finally, delayed inference implies that each frame (except near
the boundaries) appears in multiple overlapping chunks and thus
receives multiple estimates. These estimates can be aggregated,
either uniformly or with temporal weighting (e.g.,
center-weighted attention), to further smooth the output. This
strategy, shown in Algorithm~\ref{alg:averaging}, is used in
our production pipeline.

\begin{algorithm}
\caption{Inference: lookahead and averaging}
\label{alg:averaging}
\begin{algorithmic}[1]
\Require model $f$, window size $L$, delay $d$
\State Initialize buffer $X \gets [\,]$
\State Initialize list of partial estimates $H[t] \gets [\,]$
\For{each new frame with keypoints $x_t$}
	\State Append $x_t$ to $X$
	\If{length of $X > L$} \State Remove oldest frame \EndIf
	\If{length of $X = L$}
		\State $\hat{\Theta} \gets f(X)$
		\For{$i = 0$ to $L-1$}
			\State Append $\hat{\Theta}[i]$ to $H[t - L + 1 + i]$
		\EndFor
	\EndIf
	\If{$H[t-d]$ exists}
		\State Output average of $H[t-d]$
	\EndIf
\EndFor
\end{algorithmic}
\end{algorithm}

This final version balances responsiveness with stability and
avoids prediction jitter by averaging redundant estimates for
each frame. It also allows future extensions, such as weighted
averaging or temporal confidence modeling.

\section{Empirical Evaluation}
\label{sec:evaluation}

Our network is built around a transformer encoder with 6 layers,
128-dimensional feature space, and 4 heads. We train our model
using Adam optimizer for 100 epochs with a higher (0.001)
learning rate for the first 10 epochs and exponentially
decreasing lower learning rate (0.0001 -- 0.00001) for the
remaining 90 epochs. We monitor convergence on a 
validation set constituting 5\% of the data, and observe
convergence of network parameters at $\approx$80 epochs.

We train on a chunk length of 16 frames (approximately 0.5 sec).
Each batch is randomly rotated during training. These settings
are supported by ablation studies (Section~\ref{sec:ablation}).

The experiments were conducted on an NVIDIA GeForce RTX 3080 Ti GPU with 12GB of memory. 

\subsection{Real-Time Application}

We particularly target online inverse kinematics, hence inference running time and memory consumption are critical indicators to report. We measured the performance on 600 frame (20 seconds) motion sequences. The measurements are shown in Table~\ref{tbl:performance}. 

\begin{table}[H]
\centering
\caption{Performance, measured on a 20 seconds motion.}
\label{tbl:performance}
\begin{tabular}{r | c}
\toprule
GPU Memory consumption & 1.1 Gigabytes \\
Forward pass duration per chunk  & 0.029 seconds \\
Algorithm~\ref{alg:averaging} duration & 18.80 seconds \\
\bottomrule
\end{tabular}
\end{table}

Based on the observed performance, Algorithm~\ref{alg:averaging} can be run in real time upon each frame's arrival. Delay of a fraction of a second is often acceptable, and chunks can be batched, hence, with a small delay, the framework can be applied
even if resource constraints are tighter than in our setup.

\subsection{Ablation Studies}
\label{sec:ablation}

\subsubsection{Chunk Length for Training}

\begin{figure}
	\centering
	\includegraphics[width=0.7\linewidth]{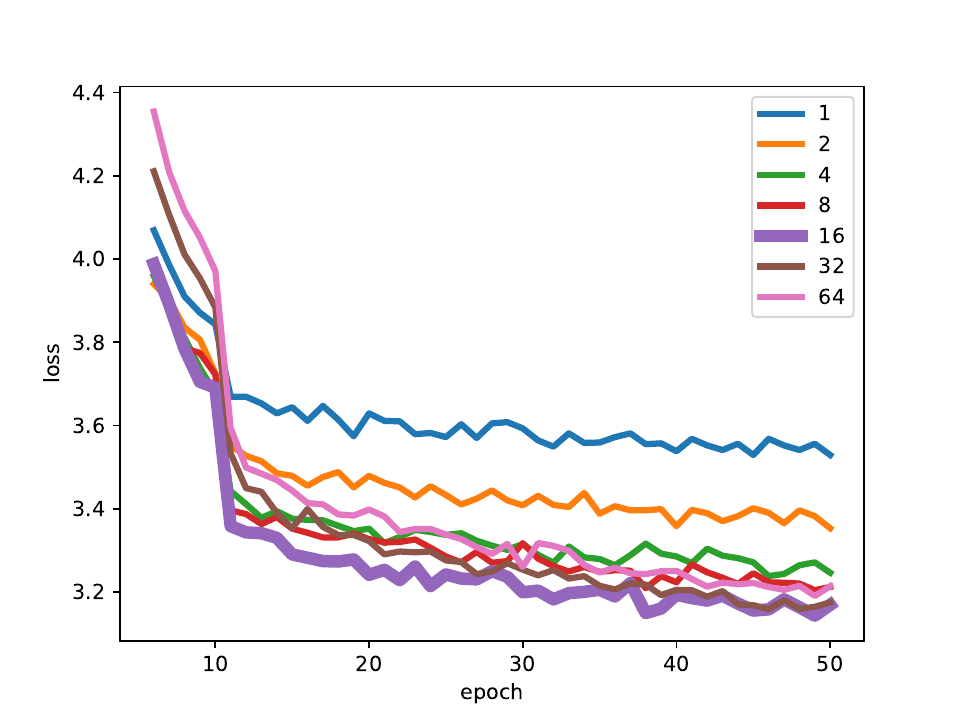}
	\caption{Loss vs. chunk length: the chunk length of 16 is sufficient
	for capturing the temporal context for inverse kinematics.}
	\label{fig:train-losses}
\end{figure}

Motions in the dataset are divided into chunks of equal length.
Too long chunks would mean high memory consumption during training
and, due to temporal position encoding, inefficient use of
the available amount of data. Chunks that are too short would
not let the network learn to capture the temporal context.
Figure~\ref{fig:train-losses} shows validation loss for
different chunk lengths. One can observe that short chunks
(the blue, orange, green, and red curves)
result in a higher loss due to insufficient temporal context.
The lowest loss over 50 training epochs is achieved for the
chunk length of 16 (the purple curve). The losses for chunks of
32 and 64 frames (the brown and pink curves) are higher due to
slower convergence and lower data efficiency.

\subsubsection{Random Rotation of Training Samples}

\begin{figure}
	\centering
	\begin{subfigure}[c]{0.49\linewidth}
		\centering
		\includegraphics[width=\linewidth]{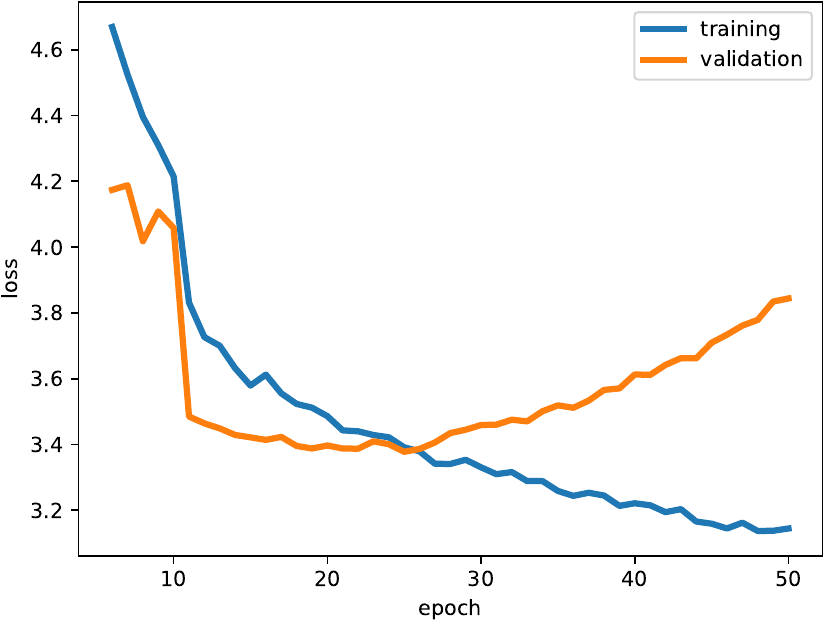}
		\caption{no rotation}
		\label{fig:no-rotation}
	\end{subfigure}
	\begin{subfigure}[c]{0.49\linewidth}
		\centering
		\includegraphics[width=\linewidth]{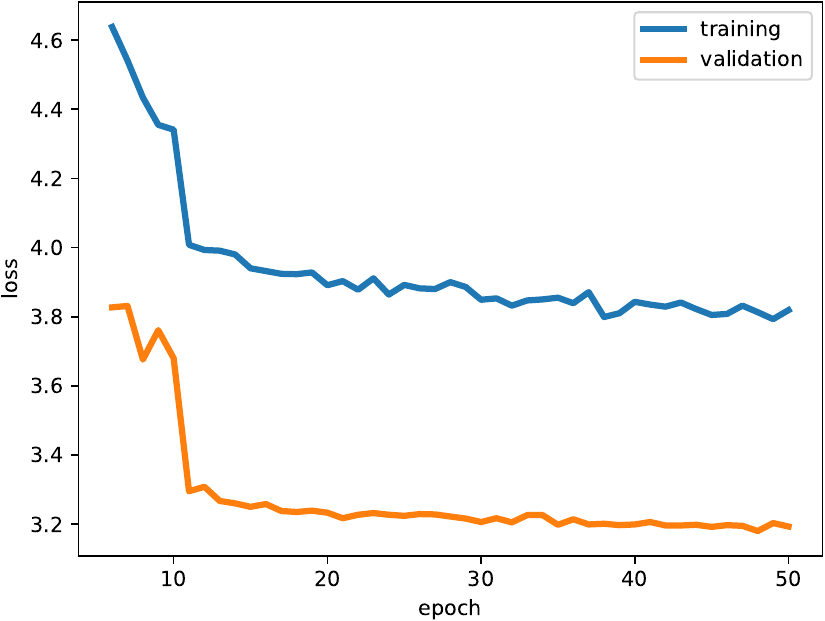}
		\caption{random rotation}
		\label{fig:rotation}
	\end{subfigure}
	\caption{Influence of random rotation of training samples
	around the Z axis:
	random rotation promotes generalization and prevents
	overfitting.}
	\label{fig:rotation-no-rotation}
\end{figure}

The network predicts rotations of all joints including the root
joint (the pelvis in our model). While rotations around the X
and Y axis are meaningful and are likely correlated with
rotations of other joints (a lying person has a different
distribution of rotations of the spine joints compared to a
standing one), rotations around the Z axis are arbitrary and
reflect the studio setup rather than properties of the motions
themselves. Training the network on motions oriented as in
the dataset causes overfitting. The network
apparently memorizes the overall motion orientations in the
horizontal plane as the cues for pose prediction
(Figure~\ref{fig:no-rotation}). To eliminate this source of
overfitting and promote generalization, we rotate each batch
of motion by a random angle between 0 and 2$\pi$ around the
Z axis and achieve proper convergence of the validation loss
without overfitting (Figure~\ref{fig:rotation}).

\subsection{Qualitative Assessment}

\begin{figure}
	\centering
	\begin{subfigure}[c]{0.39\linewidth}
		\centering
		\includegraphics[width=\linewidth]{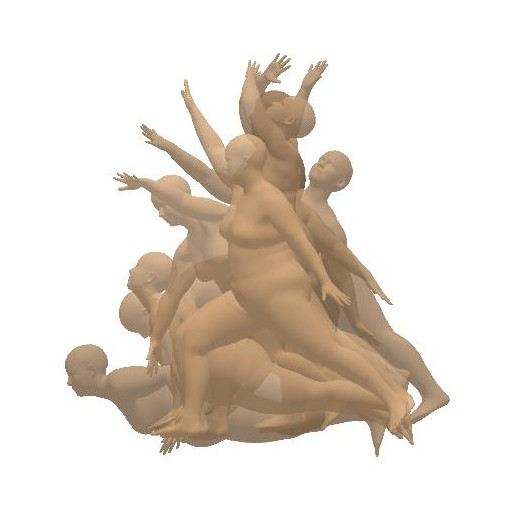}
		\caption{Yoga}
		\label{fig:yoga}
	\end{subfigure}
	\begin{subfigure}[c]{0.59\linewidth}
		\centering
		\includegraphics[width=\linewidth]{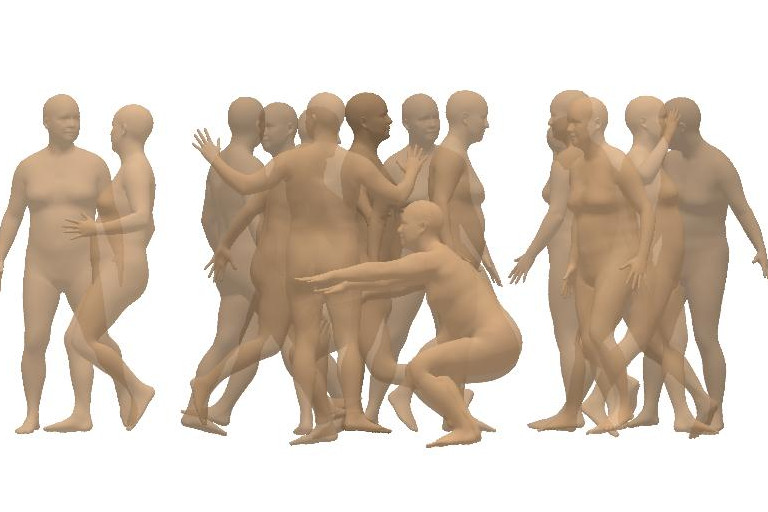}
		\caption{Walking}
		\label{fig:walking}
	\end{subfigure}
	\begin{subfigure}[c]{0.59\linewidth}
		\centering
		\includegraphics[width=\linewidth]{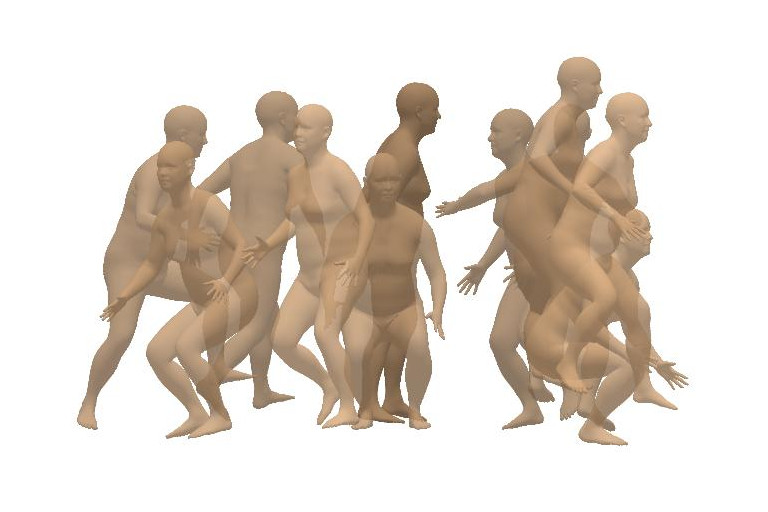}
		\caption{Jumping}
		\label{fig:jumping}
	\end{subfigure}
	\begin{subfigure}[c]{0.39\linewidth}
		\centering
		\includegraphics[width=\linewidth]{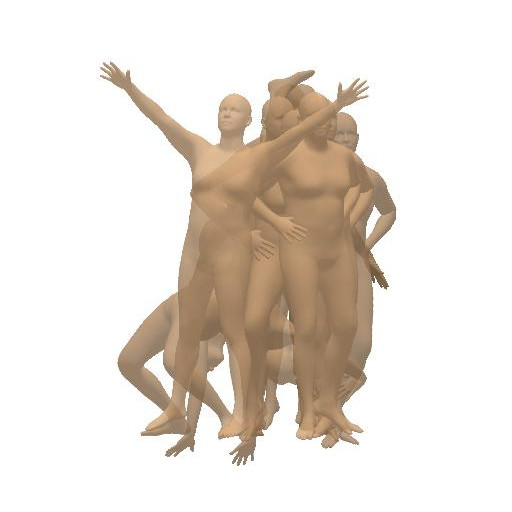}
		\caption{Dancing}
		\label{fig:dancing}
	\end{subfigure}
	\caption{Motions for qualitative assessment.}
	\label{fig:motions}
\end{figure}

To illustrate practical performance of the proposed framework on
challenging motions, we provide visual comparisons (as video
renditions) of motions reconstructed from keypoints using the
neural inverse kinematics described here and through an accurate
but slow optimization procedure, as a baseline. Motions included into the qualitative comparison are shown in Figure~\ref{fig:motions}. The videos are
published at \url{https://www.yoomtrack.com/fastneuralik}. 

\section{Conclusion}
\label{sec:conclusion}

We presented a neural inverse kinematics framework designed to balance accuracy and inference speed for real-time, markerless motion capture. The architecture, based on a transformer encoder, enables reliable online estimation of body pose and shape from 3D keypoints with minimal latency. While the current system is intentionally simple, it has proven effective and sufficient for both evaluation and deployment in production settings. Development is ongoing, and we continue to explore extensions informed by current research, including spatio-temporal attention mechanisms, robustness to occlusions, and the integration of physical priors.

\bibliographystyle{acm}
\bibliography{refs}

\end{document}

%% file: network.tex
\tikzset{
  arrow/.style={-latex,very thick},
  box/.style={draw, minimum width=2.2cm, minimum height=1cm, align=center, rounded corners=3pt, fill=blue!10},
  input/.style={draw, minimum width=2.5cm, minimum height=1cm, align=center, fill=gray!20},
  brace/.style={decorate, decoration={brace, amplitude=5pt}, thick},
  do path picture/.style={%
      path picture={%
        \pgfpointdiff{\pgfpointanchor{path picture bounding box}{south west}}%
          {\pgfpointanchor{path picture bounding box}{north east}}%
        \pgfgetlastxy\x\y%
        \tikzset{x=\x/2,y=\y/2}%
        #1
      }
  },
  sin wave/.style={do path picture={    
  	\draw [line cap=round] (-3/4,0)
  	sin (-3/8,1/2) cos (0,0) sin (3/8,-1/2) cos (3/4,0);
  	}
  }
}

\begin{tikzpicture}[node distance=1.0cm and 0.6cm, very thick]

  \node[input] (input) {3D keypoints\\ (B $\times$ T $\times$ 3$\cdot$K)};
  
  \node[box, below=of input] (readin) {Linear\\ read-in};

  \node[circle, draw, minimum size=0.3cm, inner sep=0pt, right=of readin] (addposenc) {\textbf{+}};
  \node[circle, draw, sin wave, minimum size=0.75cm, below=0.5cm of addposenc] (posenc) {};
  \node[minimum width=2.2cm, minimum height=1cm, align=center,below=0.1cm of posenc] () {Positional\\ encoding};

  \node[box, right=of addposenc] (transformer) {Transformer\\ encoder};

  \node[draw=none, right=0.6cm of transformer] (outseq) {};

  \node[box, above=of outseq] (attnreadout) {Attention\\ pooling};
  \node[box, below=of outseq] (mlpreadout) {MLP\\ read-out};

  \node[input, right=of attnreadout] (shape) {Shape};
  \node[input, right=of mlpreadout, yshift=0.5cm] (trans) {Root\\ translation};
  \node[input, right=of mlpreadout, yshift=-0.5cm] (rots) {Joint\\ rotations};

  \draw[arrow] (input) -- (readin);
  \draw[arrow] (readin) -- (addposenc);
  \draw[arrow] (posenc) -- (addposenc);
  \draw[arrow] (addposenc) -- (transformer);
  \draw[arrow] (transformer) -- (attnreadout);
  \draw[arrow] (transformer) -- (mlpreadout);
  \draw[arrow] (attnreadout) -- (shape);
  \draw[arrow] (mlpreadout) -- (trans);
  \draw[arrow] (mlpreadout) -- (rots);

  \draw[brace] ([xshift=-0.55cm,yshift=0.35cm]transformer.north east) -- ([xshift=-0.2cm,yshift=0.35cm]attnreadout.west);
  \node[below left=-0.05cm and 0.35cm of attnreadout.west] {\footnotesize \textbf{over all frames}};

\end{tikzpicture}